\documentclass[final]{cvpr}
\usepackage{nopageno}

\usepackage{times}
\usepackage{epsfig}
\usepackage{graphicx}
\usepackage{amsmath}
\usepackage{amssymb}
\usepackage{stmaryrd}
\usepackage{tabularx,booktabs}

\usepackage[pagebackref=true,breaklinks=true,colorlinks,bookmarks=false]{hyperref}

\def\cvprPaperID{2} % *** Enter the CVPR Paper ID here
\def\confYear{CVPR 2021}

\begin{document}

%%%%%%%%% TITLE
% \title{Deep Inverse Makeup Graphics for Real Time Video Makeup Transfer}
\title{Deep Graphics Encoder for Real-Time Video Makeup Synthesis from Example}


\author{Robin Kips$^{1, 2}$,
% L'Or\'eal Research and Innovation\\
% France\\
% {\tt\small robin.kips@loreal.com}
% \and
Ruowei Jiang$^3$,
% Modiface\\
% Canada\\
% {\tt\small irene@modiface.com }
% \and
Sileye Ba$^1$,
% L'Or\'eal Research and Innovation\\
% France\\
% {\tt\small sileye.ba@loreal.com }
% \and
Edmund Phung$^3$,
% Modiface \\
% Canada \\
% {\tt\small edmund@modiface.com }
% \and
Parham Aarabi$^3$,
% Modiface \\
% Canada \\
% {\tt\small parham@modiface.com}
% \and
Pietro Gori$^2$ \\
% LTCI, T\'el\'ecom Paris, Institut Polytechnique de Paris \\
% France \\
% {\tt\small pietro.gori@telecom-paris.fr}
% \and
Matthieu Perrot$^1$,
% L'Or\'eal Research and Innovation\\
% France \\
% {\tt\small matthieu.perrot@loreal.com}
% \and
Isabelle Bloch$^4$,
% Sorbonne Universit\'e, CNRS, LIP6, Paris, France\\
% France \\
% {\tt\small isabelle.bloch@telecom-paris.fr}
\and
$^1$ L'Or\'eal Research and Innovation, France\\
% \and
$^2$ LTCI, T\'el\'ecom Paris, Institut Polytechnique de Paris, France, 
% \and
$^3$ Modiface, Canada\\
% \and 
$^4$ Sorbonne Universit\'e, CNRS, LIP6, Paris, France\\
{\tt\small \{robin.kips, sileye.ba, matthieu.perrot\}@loreal.com, \{irene,edmund,parham\}@modiface.com} \\
{\tt\small pietro.gori@telecom-paris.fr, isabelle.bloch@sorbonne-universite.fr}
%{\tt\small \{irene,edmund,parham\}@modiface.com}
}

\maketitle

%%%%%%%%% ABSTRACT
\begin{abstract}
While makeup virtual-try-on is now widespread, parametrizing a computer graphics rendering engine for synthesizing images of a given cosmetics product remains a challenging task. In this paper, we introduce an inverse computer graphics method for automatic makeup synthesis from a reference image, by learning a model that maps an example portrait image with makeup to the space of rendering parameters. This method can be used by artists to automatically create realistic virtual cosmetics image samples, or by consumers, to virtually try-on a makeup extracted from their favorite reference image.
\end{abstract}

%%%%%%%%% Introduction
\section{Introduction}
%[look into parametric design keyword/ creative design / creative AI] [add a graphics code definition : the set of parameters the control the makeup material, different from scene parameters].
Virtual-try-on technologies are now largely spread across online retail platforms and social media. In particular, for makeup, consumers are able to virtually try-on cosmetics in augmented reality before purchase. While creating virtual makeup for entertaining purposes is easy, parametrizing a rendering engine for synthesizing realistic images of a given cosmetics remains a tedious task, and requires expert knowledge in computer graphics. Furthermore, consumers are often prompted to select among a set of predefined makeup shades, but they cannot try makeup look from a reference inspirational images on social media. 

\begin{figure}[t!]
\begin{center}
   \includegraphics[width=1.0\linewidth]{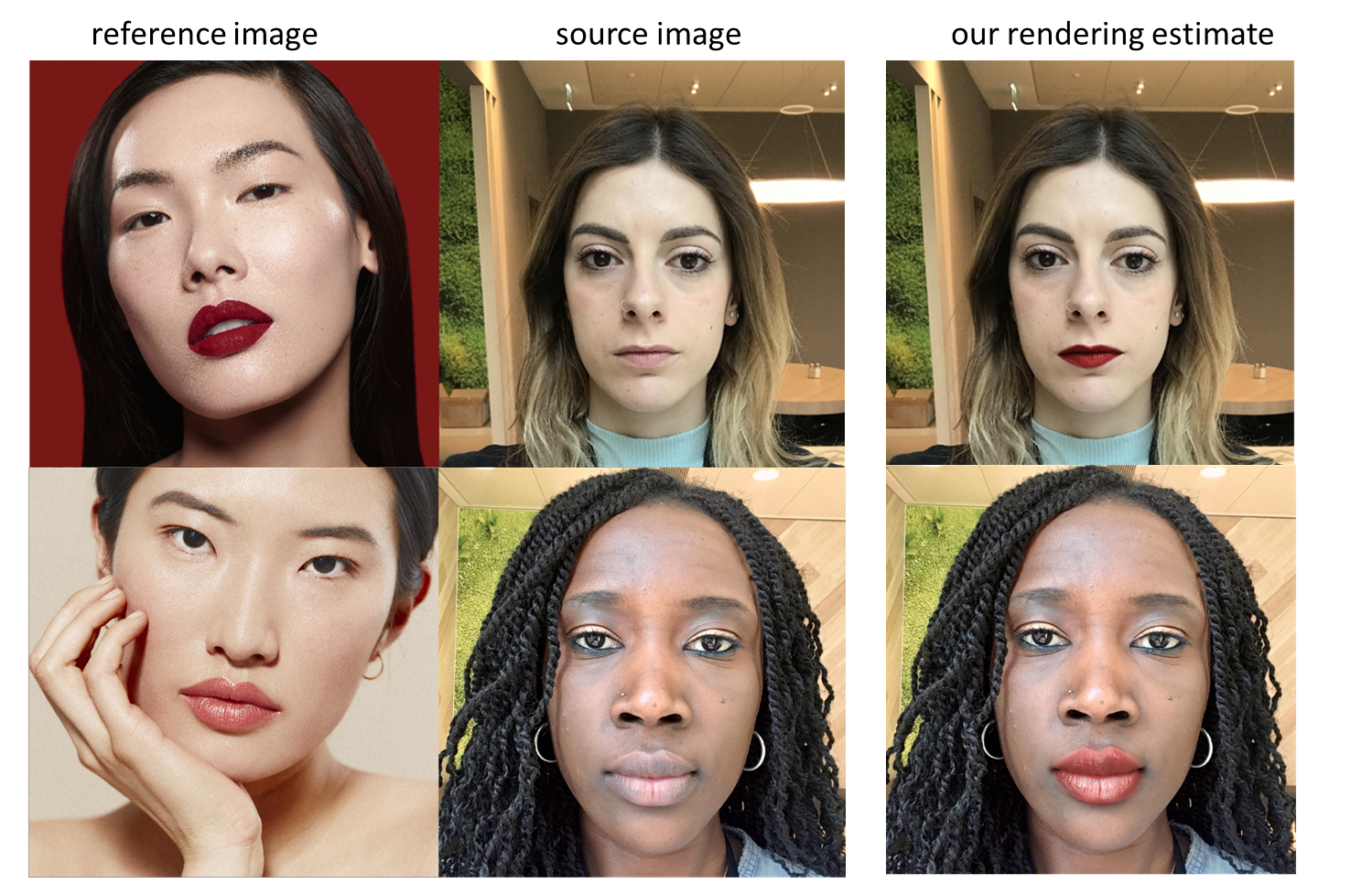}
\end{center}
   \caption{Examples of lipstick transfer from example images using our proposed method.}
\label{fig:armani_example}
\end{figure}

In the past few years, the field of computer vision attempted to provide a solution to this problem through advances in the \textit{makeup style transfer} task. This task consists in extracting a makeup style from a reference portrait image, and synthesizing it on the target image of a different person. State-of-the-art methods for this task \cite{kips2020gan,li2018beautygan} are based on a similar principle. First, makeup attributes are extracted using a neural network and represented in a latent space. Then, this neural makeup representation is decoded and rendered on the source image using a generative model, such as GAN \cite{goodfellow2014generative} or VAE \cite{kingma2014auto}. 

\begin{figure*}[t!]
\begin{center}
   \includegraphics[width=0.7\linewidth]{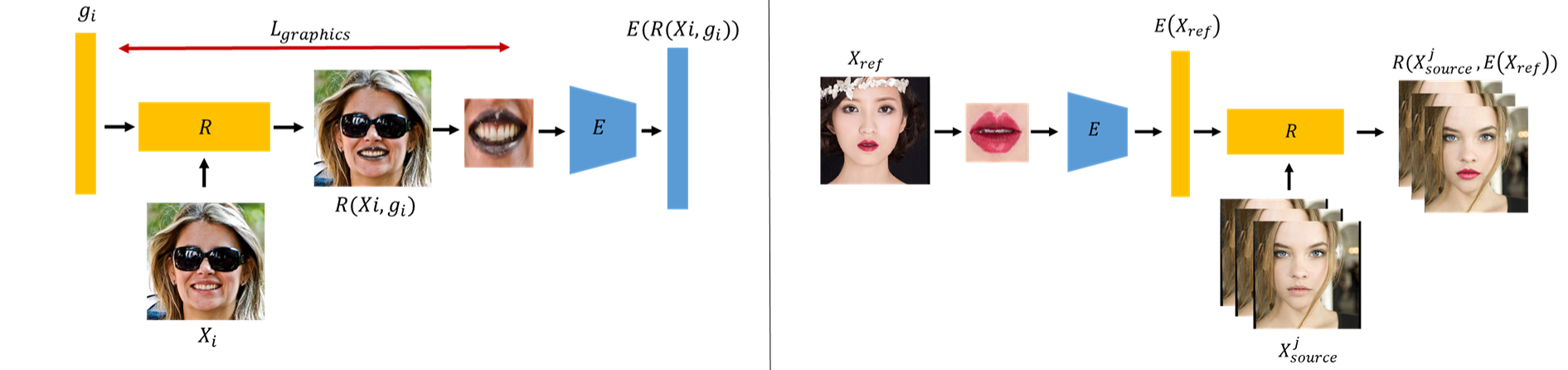}
\end{center}
   \caption{Left: training procedure of our model. We sample a graphics parameters vector $g_i$ and render a corresponding image using a renderer $R$ and a random source image $X_i$. Then, the inverse graphics encoder $E$ is trained to map the image to the space of graphics parameters with minimum error. Right: inference pipeline. A reference image $X_{ref}$ is passed to the inverse graphics encoder to estimate the corresponding makeup graphics parameters. Then this code can be used as input to the rendering engine, to render the reference makeup on videos in real-time. To facilitate training and increase the proportion of relevant pixels in the image, $E$ is trained on crops of eyes and lips.}
\label{fig:model}
\end{figure*}

%The use of generative models, such as GAN \cite{goodfellow2014generative} or VAE \cite{kingma2014auto}, for rendering-based tasks allows producing realistic results but suffers from various limitations in practice.

The use of generative model for addressing rendering-based problem, also denoted as \textit{neural rendering} \cite{tewari2020state}, allows producing realistic results but suffers from various limitations in practice. First, such approaches are currently impossible to use for real-time inference on portable devices. Furthermore, training generative models for video data is an emerging field, and even state-of-the-art models produce sequences of frames with noticeable temporal inconsistencies \cite{chu2020learning, thimonier2021styletemp}. 
%there is no clear strategy on how to avoid temporal inconsistency when using framed-based generative models on video inference \SB{Why not GANs for videos generation}.
Finally, generative methods are highly dependent on the training data distribution and might fail in the case of extreme examples, such as unusual makeup colors. These drawbacks make the use of current makeup transfer methods unusable in practice for consumer virtual try-on applications. 

On the other hand, computer graphics methods offer symmetric advantages. Even though the most advanced rendering techniques require intensive computations, many graphics-based methods can be used to produce realistic images in real-time, even on portable devices. As opposed to generative methods, graphics-based methods do not rely on training data, and can render videos without time inconsistency issues.
%They can be used for real-time video inference on portable devices and do not rely on training data \SB{Dont they have the same problem for realtime inference, efficient rendering needs GPUs}. 
However, they need to be carefully parametrized to render a given cosmetic product in a realistic manner. In practice, this is a tedious work that requires expert knowledge in computer graphics.

Recent works on \textit{inverse rendering} introduced methods for estimating graphics attributes using differentiable rendering \cite{kato2020differentiable}. Such methods \cite{gao2019deep} propose to estimate parameters such as shape or BRDF by computing a forward rendering using an engine with differentiable operations, associated with gradient descent to optimize the graphics attributes with respect to one or multiple example images. However, this class of problem is often ill-posed, attempting to compute high-dimensional BRDF from RGB images. Furthermore, most real-time renderers are not differentiable in practice, and would require costly changes in computer graphics methods or implementation choices. To the best of our knowledge, there is no previous work in inverse computer graphics for makeup.

In this paper, we introduce a novel method based on deep inverse computer graphics for automatically extracting the makeup appearance from an example image, and render it in real-time, for a realistic virtual try-on on portable devices. Examples of our results are illustrated in Figure \ref{fig:armani_example}. Our contributions can be summarized as follows:

\begin{itemize}
    \item We introduce a simple but powerful framework for learning an inverse graphics encoder network that learns to map an image into the parameter space of a rendering engine, as described in Figure \ref{fig:model}. This is a more efficient and compact approach than inverse rendering, and does not require the use of a differentiable renderer. 

    \item We demonstrate the effectiveness of this framework for the task of makeup transfer, outperforming state-of-the-art results, and achieving high resolution real-time inference on portable devices.
\end{itemize}

% - \cite{kulkarni2015deep} auto encoder using graphics code in the latent space. 
% - \cite{gao2019deep} : SVBRDF estimation. Perform optimization in the latent space instead of SVBRFD space (more compact. 

% - inverse rendering required to optimize parameters like camera, light, shape, to compute a pixel wise loss. In our case the objective is to recover the graphics code.
% in our case not really inverse rendering because no forward rendering step in the training pipeline. 

% Why not formulate as a SVBRDF estimation method ?
% - SVBRDF not relevant for makeup ? 
% - SVBRDF is ill-defined. For our application we can work in a simplified space. 
% - object specialized vs general material properties extraction.
% - svbrdf simplified ?

\section{Method}

%In this section, we describe the various components of our deep inverse graphics method. 
\subsection{Computer graphics makeup renderer} \label{cg_renderer}
\begin{figure*}
    \centering
    \includegraphics[width=0.65\textwidth]{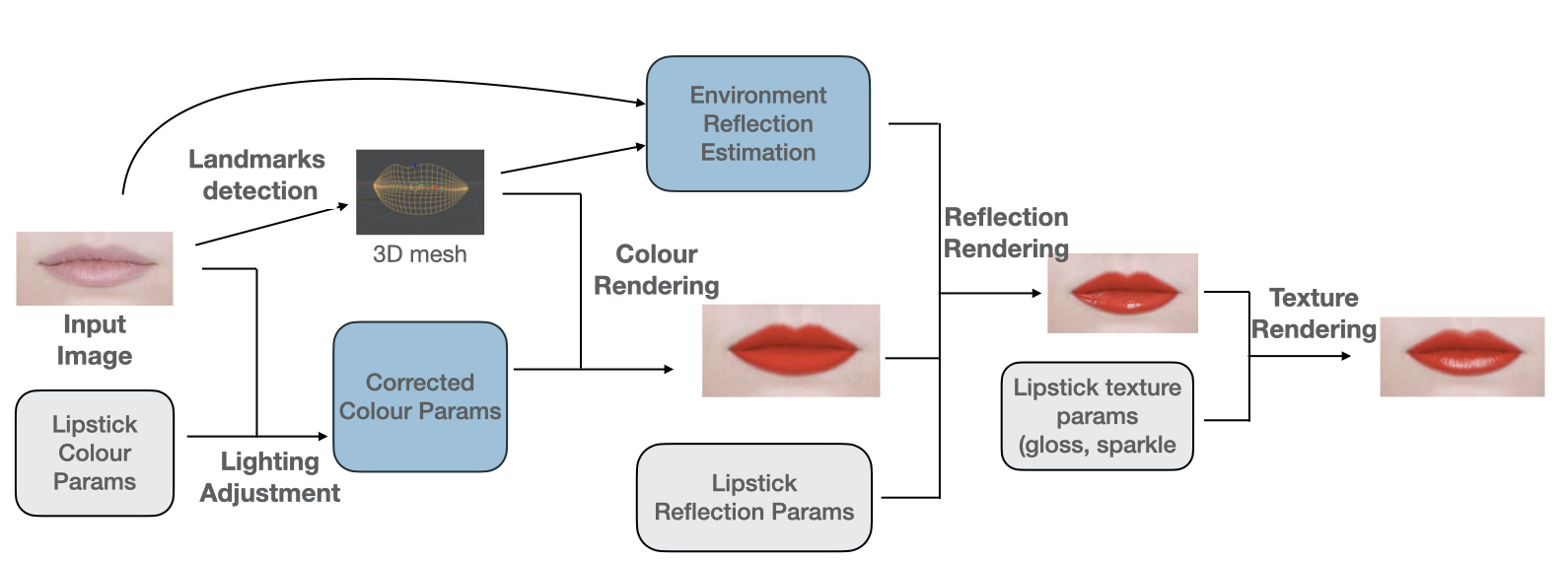}
    \caption{Our computer graphics rendering pipeline. While the makeup parameters are fixed prior to rendering, some parameters such as the lip mesh and the illuminant are estimated on each frame to render.}
    %\PG{it's too small... either you put it on two columns or you remove it. I think}} [I will see after adding the plot on eyes, otherwise it will be moved to supplementary.]
    \label{fig:rendering_pipeline}
\end{figure*}

To achieve a realistic makeup rendering, we use a graphics-based system, that considers rendering parameters of color and texture features of any given cosmetics, such as described in~\cite{li2019lightweight}. Figure \ref{fig:rendering_pipeline} illustrates the complete rendering pipeline for lipstick simulation. 
%To achieve a realistic rendering, we design a graphics-based lipstick rendering system that considers rendering parameters of color and texture features of any given lipstick. In Table \ref{tab:param_desc}, we describe the major rendering parameters.

% \begin{table}[h!]
%     \centering
%     \caption{Descriptions of major rendering parameters. The complete rendering engine includes a total of 17 parameters. }
%     \begin{tabular}{ccc}
%         \toprule
%         Name  &  Description & Range \\
%         \midrule
%         intensity &makup opacity. & [0, 1]\\
%         R,G,B,A & makeup colour. & [0, 256) $\in \mathbb{N}$ \\
%         gloss&The amount of gloss on the lips& [0, inf), usually less than 4\\
%         glossDetail & Controls the size of the bright spots in the gloss & [0, 1]\\
%         envMappingIntensity & The opacity of the environment map reflection. & [0, 1]\\
%         \bottomrule
%     \end{tabular}
%     \label{tab:param_desc}
% \end{table}
\begin{table}[h!]
    \centering
    \caption{Descriptions of rendering parameters used in our graphics parameters vector representing the makeup material. The complete rendering engine includes a total of 17 parameters. }
    \small
    \begin{tabular}{cc}
        \toprule
        Description & Range \\
        \midrule
        Makeup opacity & [0, 1]\\
        R,G,B & $\llbracket 0, 255 \rrbracket$ \\
        Amount of gloss on the makeup & $[0, +\infty)$ \\
        Gloss Roughness & [0, 1]\\
        Reflection intensity & [0, 1]\\
        \bottomrule
    \end{tabular}
    \label{tab:param_desc}
    
\end{table}

To obtain real-time inference, we first estimate a 3D lips mesh from estimated 3D facial landmarks. The rendering is then completed in two parts: 1) lips recoloring;  2) a two-step texture rendering based on environment reflection estimation and other controlled parameters. A similar rendering pipeline also applies to other makeup products. Table~\ref{tab:param_desc} describes the major rendering parameters used for representing a makeup product.

\subsection{Inverse Graphics Encoder}
We introduce a simple yet powerful framework to train an inverse graphics encoder that learns to project an example image to the parameter space of a rendering engine. In the case of makeup image synthesis, this allows us to automatically compute the renderer parametrization in order to synthesize a makeup similar to that of a reference example image.  

The training procedure of our framework is described in Figure~\ref{fig:model}. We denote by $R$ the computer graphics rendering engine, taking as input a portrait image $X$ and parametrized by  $g$, the vector of parameters representing the makeup material that we name the \textit{graphics parameters}. Each component of $g$ is described in Table~\ref{tab:param_desc}. Our objective is to train an encoder $E$, so that given an example makeup image $X_{ref}$, we can estimate the corresponding graphics parameters $\hat{g} = E(X_{ref})$ to render images with the same makeup appearance. 

Since the renderer $R$ is not differentiable, we do not use the \textit{inverse rendering} approach and propose to learn $E$ through optimization in the space of graphics parameters. This is a more compact problem than inverse rendering or material appearance extraction, and does not require a time-consuming gradient descent step for inference. Instead, we train a special-purpose machine learning model that learns an accurate solution for a given renderer and graphics parameters choice.
Mathematically, we denote by $g_i$ a randomly sampled graphics parameters vector, and $X_i$ a random portrait image. Thus, our model $E$ is trained to minimize the following objective function :
% $$L_{graphics} = 1/n$$
%$$ L_{graphics} = \frac{1}{n}  \sum_{i=1}^n \left\lVert g_i - E(R(X_i,g_i)) \right\rVert ^2_2$$
$$ L_{graphics} = \frac{1}{n}  \sum_{i=1}^n \left\lVert g_i - E(R(X_i,g_i)) \right\rVert ^2$$

Our approach does not depend on a training dataset of natural images, but only on a sampling of graphics parameters that we control entirely at training time. Therefore, in comparison to existing methods, our model is not sensitive to bias in available training datasets. 
Instead, we can select a graphics parameters distribution that samples the entire space of rendering parameters, which leads to better performance specially in cases of rare and extreme makeup examples. In our experiments, we used an EfficientNet B4 architecture \cite{tan2019efficientnet} to represent $E$, replacing the classification layer by a dense ReLU layer of the same dimension as $g$. This light architecture is chosen in order to obtain a portable model with real-time inference. We construct two synthetic datasets by sampling $n=15000$ graphics parameters vectors for eyes and lips makeup, and rendering them on random portrait images from the \textit{ffhq} dataset~\cite{karras2019style}, using the renderer described in Section \ref{cg_renderer}. 
%In order to obtain a data distribution that is both diverse and realistic, the graphics code are sampled using a mixture of uniform distribution, and a multivariate gaussian distribution fitted on real rendering parameters set by makeup experts.
To obtain a realistic data distribution, the graphics parameters are sampled using a multivariate normal distribution fitted on real rendering parameters set by makeup experts to simulate real cosmetics. Furthermore, we also sample graphics parameters using a uniform distribution, in order to reinforce data diversity and improve our model performance on extreme makeup examples. 
%\PG{Here, you should give more details. Not so clear how you use uniform distribution and/or Gaussian distributions.} 
Finally, our model is trained on crops of lips and eyes in order to increase the proportion of relevant pixels in the training image. %\PG{So you use a crop during training and a full facial image during test ?}

At the inference time, an example makeup image is passed to our inverse graphics encoder to estimate a corresponding graphics parameters vector. These parameters can then be used to render the extracted makeup attributes on any source video in real-time using $R$. Since the graphics parameters are fixed for each makeup, the inverse graphics encoder only needs to be run once per reference makeup image, and can then be used to render later video for any consumer. The inference pipeline is illustrated in Figure~\ref{fig:model}.

%Finally, our method is trained on rendered images only, and therefore relies on the assumption that $R$ produces realistic results, leading to a reduced domain gap on natural images at inference time. %\PG{what do you means with synthetic ? Is it an image of a person also during training non ?}

\section{Experiments and Results}

%In the following section, we describe various experiments to illustrate the performance of our framework. 

\subsection{Qualitative experiments}

\begin{figure*}[t!]
\begin{center}
   \includegraphics[width=0.9\linewidth]{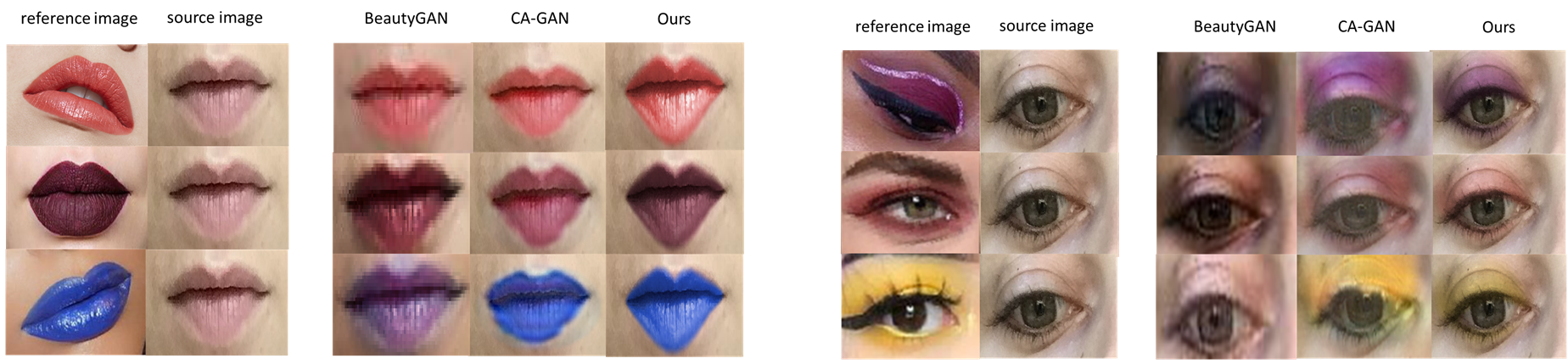}
\end{center}
   \caption{Qualitative comparison on lipstick and eye-shadow synthesis against state of the art makeup transfer methods. Our method is capable of reproducing realistic rendering in high resolution for makeup with various colors and textures. The eye-shadow application zone and intensity are not part of the estimated graphics parameters, but set by the user at rendering time according to their preferences.
   }
\label{fig:quali_lips}
\end{figure*}

In order to obtain a qualitative evaluation of our framework, we compare our approach to two state-of-the-art methods of makeup transfer: BeautyGAN~\cite{li2018beautygan} and CA-GAN~\cite{kips2020gan}. 
We extract makeup from multiple reference images with various colors and glossiness levels, and synthesize makeup on the same source image, as illustrated in Figure~\ref{fig:quali_lips}. Our method produces more realistic results than existing makeup transfer methods, and is capable of accurately rendering the appearance of lipsticks and eye-shadow with various colors and textures, without any loss of image resolution. Furthermore, since our method is not dependent on the distribution of a training dataset, it largely outperforms other methods on extreme makeups such as blue lipstick or yellow eye-shadow, as shown in Figure~\ref{fig:quali_lips}. 

In our problem formulation, the eye-shadow application zone and intensity are not part of the estimated graphics parameters, but set by the user at rendering time according to their preferences. This choice allows for an increased user control on the makeup style, at the cost of not reproducing automatically the entire eye makeup style of the reference image.
Finally, to give the reader more insight about our model, we provide example videos as supplementary materials, as well as an interactive demo application for lipstick transfer.%\footnote{also accessible at \url{http://ec2-3-224-169-8.compute-1.amazonaws.com/copy_param_demo}}

\subsection{Quantitative experiments}

In order to compare our results with existing methods, we reproduce the quantitative evaluation of makeup transfer performance on lipstick, introduced in \cite{kips2020gan}. More precisely, we use the dataset provided by the authors with 300 triplets of reference portraits with lipstick, source portraits without makeup, and associated ground-truth images of the same person with the reference lipstick. We compute the accuracy of our model over various perceptual metrics %\IB{il faudrait les preciser ?} %RK : malheureusement je ne peux pas citer les 2 papiers en question car les références comptent dans les 4 pages... Je renvoie donc vers le papier qui lui rentre dans le detail de ces métriques. 
and report the results in Table~\ref{tab:quant_mu_transfer}. These results confirm that our framework outperforms the existing makeup transfer methods. %\PG{Here, maybe in the next paper, it would be interesting to have paired images (with/without lipstick) et avec les même conditons de photo} \PG{Question, si tu remets l'image d'output comme entrèe dansn E, est-ce que tu obtiens le g utilisé pour la construire ?  }

\begin{table}[t]
\small
 \caption{Quantitative evaluation of the makeup transfer performance using a dataset of groundtruth triplet images.}
\centering
\begin{tabular}{c c c c}
\hline
 Model & L1 & 1-MSSIM \cite{wang2003multiscale} & LPIPS \cite{zhang2018unreasonable}\\
\hline
BeautyGAN \cite{li2018beautygan} & 0.123 & 0.371 & 0.093\\
\hline
CA-GAN \cite{kips2020gan}& 0.085 & 0.304 & 0.077\\
\hline
Ours & \textbf{0.083} & \textbf{0.283} & \textbf{0.060}\\
\hline
\end{tabular}
\label{tab:quant_mu_transfer}
\end{table}

\subsection{Inference Speed}

An important limitation of generative-based methods for makeup transfer is their inference speed with limited resources, especially on mobile platforms. For instance, the StarGAN \cite{choi2018stargan} architecture used in CA-GAN takes 18 seconds to synthesize a 256x256 image on an Ipad Pro with a A10X CPU. Even though some optimization is possible using GPU or neural inference special-purpose chips, this makes the use of generative models currently prohibitive for real-time consumer applications.

In comparison, our method uses a neural network not on every frame of the source video, but only once to compute the graphics parameters vector sent to the renderer. Furthermore, our graphics encoder is based on EfficientNet-lite-4 \cite{tan2019efficientnet}, an architecture adapted to mobile inference, reportedly reaching an inference time of 30ms per image on a Pixel 4 CPU \cite{effnetliteblog}. Thus, the additional computational time introduced by our graphics encoder can be considered negligible when generating a video. To illustrate the inference time of our video solution, we profile our computer graphics pipeline on different mobile devices. We use the landmarks detection model described in \cite{li2019lightweight} and convert it to NCNN \cite{ncnn2018framework} 
%\IB{pas defini ?}
to make it runnable on mobile platforms. To get accurate profiling results, we skip the first 100 frames and average the results of the next 500 frames for each device. As shown in Table \ref{tab:profiling}, our system is able to achieve excellent performance even on old devices such as Galaxy S7.

\begin{table}[h]
\small
    \centering
    \caption{Profiling of our graphics rendering pipeline on 3 different devices. Since our graphics encoder is only used once before the rendering and not at each frame, we consider its time is negligible in the video synthesis.
    }
    \begin{tabular}{p{21mm}ccc}
        \toprule
        Device  &  Detection  & Rendering  & Display Time \\
        \midrule
        Galaxy S21 & 11.97ms & 14.95ms &2.12ms\\
        Pixel 4 & 18.32ms & 19.54ms &2.23ms\\
        Galaxy S7 & 27.89ms & 58.55ms &10.68ms\\
        \bottomrule
    \end{tabular}
    \label{tab:profiling}
\end{table}

\section{Conclusion}
We introduced a method for learning a special-purpose inverse graphics encoder that maps an example image to the space of a renderer parameters, even in the absence of a differentiable renderer.
We showed that our framework can be applied to the task of makeup transfer, allowing non-expert users to automatically parametrize a renderer to reproduce an example makeup.
In the future, we intend to improve our framework with a larger definition of the graphics parameters, such as including the estimation of the eye makeup application region.  

%- no need for differentiable renderer.
%- improvements : add shape parameters for the eye makeup, other task with the encoder (product recommendation, train a better generator, ...)
%- [pespective of validation stud ]

{\small
\bibliographystyle{ieee_fullname}
\bibliography{egbib}
}

\end{document}

%%%%%%%%% TITLE
% \title{Deep Inverse Makeup Graphics for Real Time Video Makeup Transfer}
\title{Supplementary materials : Deep Graphics Encoder for Real Time Video Makeup Synthesis from example}

% \author{Anonymous submission}

\maketitle

%%%%%%%%% ABSTRACT
% \onecolumn

\section{Interactive Demo}

 To illustrate the performance of our framework, we share an interactive demo for automatic lipstick synthesis from example, available at~\url{http://ec2-3-224-169-8.compute-1.amazonaws.com/copy_param_demo}. Given an example reference image uploaded by the user, we estimate the corresponding lipstick graphics parameters using our deep graphics encoder and display them on screen. Then, we render the estimated lipstick on an example portrait image. 
 However notice that this interactive application is hosted on a server for demonstration purposes, and does not reflect the inference speed of the real-time pipeline that needs to run locally for better performance.
 
 \section{Example videos}
We provide example videos of makeup synthesis from example images using our framework on typical virtual-try-on videos. These videos are accessible at the following url : 

\begin{itemize}
  \item \url{https://youtu.be/GmciY9rUMOw} for lipstick synthesis.
  \item \url{https://youtu.be/0dMrf0yZvUw} for eye-shadow synthesis. 
\end{itemize}
% - \url{https://youtu.be/GmciY9rUMOw} for lipstick synthesis.
% - \url{https://vimeo.com/528224882} for eye-shadow synthesis. 

\section{Training graphics parameter distribution}

\begin{figure*}[!]
\begin{center}
   \includegraphics[width=1.0\linewidth]{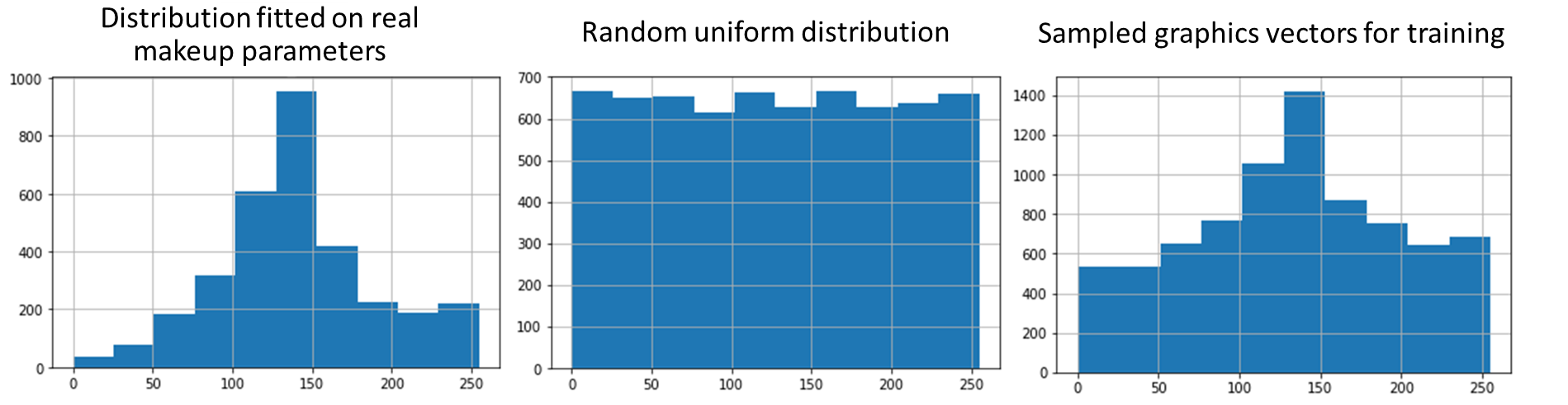}
\end{center}
   \caption{The training distribution of graphic parameters is a mixture of uniform distribution for diversity, and a distribution fitted on real makeup data for realism.
   }
\label{fig:distrib_params}
\end{figure*}

A particularity of our framework is that we fully control the distribution of graphics vector used to train our deep graphics encoder. Thus, we propose to sample the training graphics vector from a distribution that is both realistic and diverse. To obtain a realistic data distribution, the graphics parameters are sampled using a multivariate normal distribution fitted on real rendering parameters set by makeup experts to simulate real cosmetics. Furthermore, we also sample graphics parameters using a uniform distribution, in order to reinforce data diversity and improve our model performance on extreme makeup examples. Each of these distributions are illustrated in figure~\ref{fig:distrib_params}. 

% \section{Quantitative evaluation}

% In order to reinforce the quantitative evaluation of our model, we present in table

% \begin{table}
%  \caption{Test set metrics of our lipstick graphics encoder }
% \centering
% \begin{tabular}{c c c}
% \hline
%  Gaphics Parameter & MAE & variable range\\
% \hline
% color_r &  28.25 & $[0, 255]$ \\
% \hline
% color_g &  89.93 & $[0, 255]$ \\
% \hline
% color_b &  83.64 & $[0, 255]$ \\
% \hline
% gamma & 1.42 & $[0, +\infty)$\\
% \hline
% wetness & 0.29 & [0, 1]\\
% \hline
% glossDetail & 0.29 & [0, 1] \\
% \end{tabular}
% \label{tab:quant_mu_transfer}
% \end{table}

\section{Additional Qualitative examples}
\begin{figure*}[!]%[!]
\begin{center}
   \includegraphics[width=1.0\linewidth]{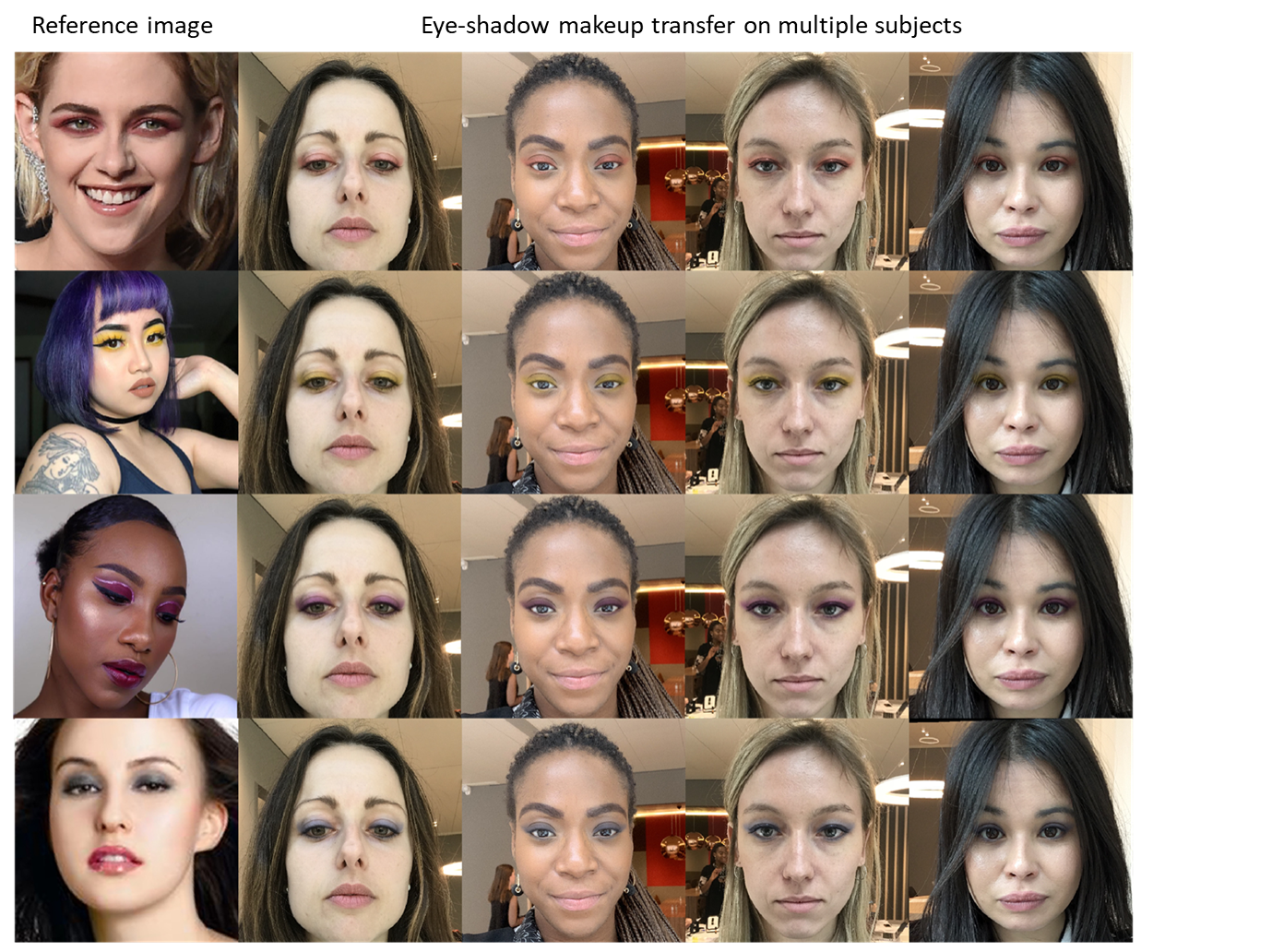}
\end{center}
   \caption{Eye-shadow synthesis from example image}
\label{fig:ff_lips}
\end{figure*}

\begin{figure*}[!]%[t!]
\begin{center}
   \includegraphics[width=1.0\linewidth]{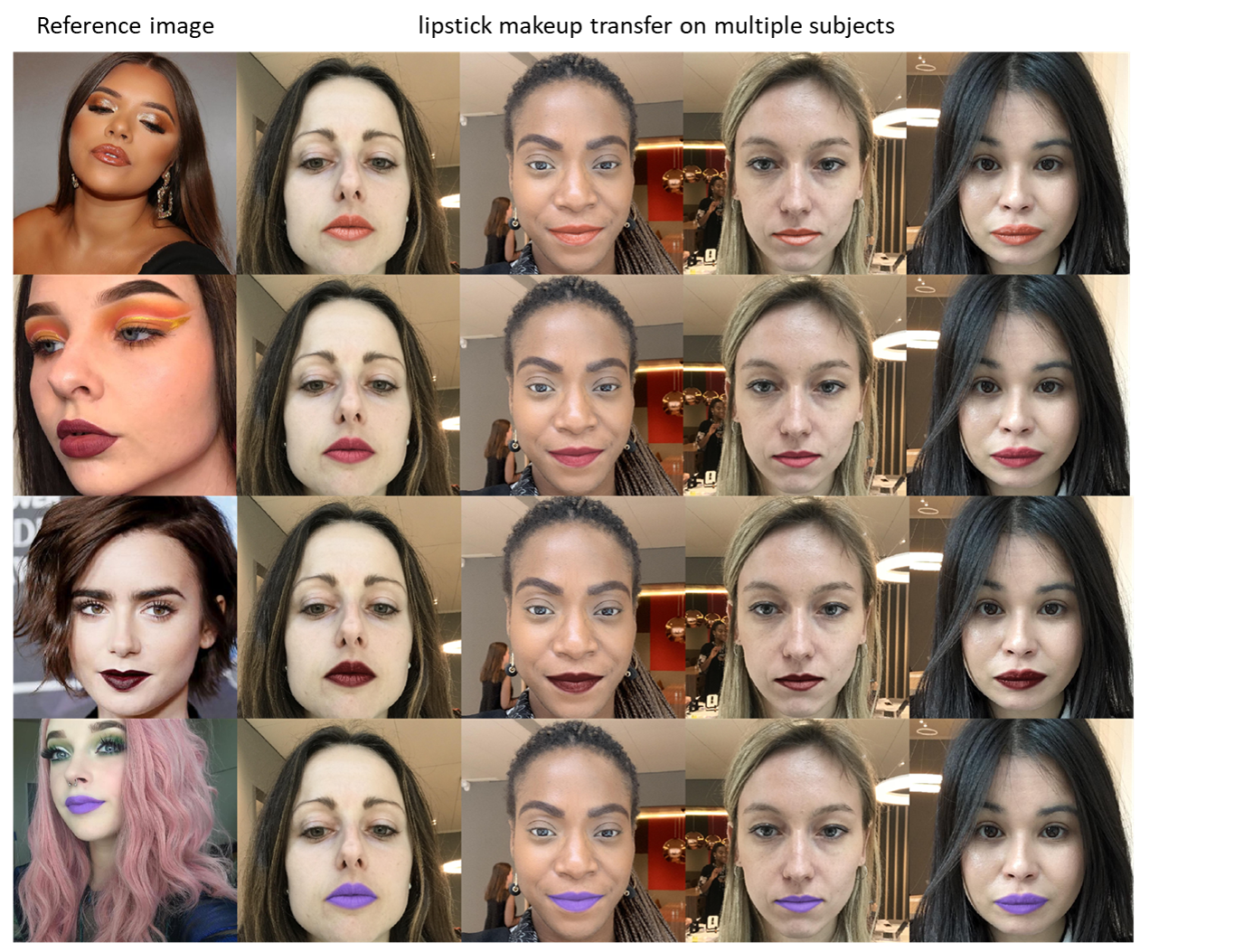}
\end{center}
   \caption{Lipstick synthesis from example image}
\label{fig:ff_eeys}
\end{figure*}

In this section we display additional examples of the performance of our framework for lipstick and eye-shadow synthesis from example images. Figures \ref{fig:ff_lips} and \ref{fig:ff_eeys} presents these results on panelists with various skin tones.


